\documentclass[sigconf, authorversion]{acmart}

\usepackage{booktabs} 

\usepackage{booktabs} 
\usepackage[utf8]{inputenc}
\usepackage[T1]{fontenc}
\usepackage{algorithmic}
\usepackage[ruled,vlined]{algorithm2e}
\usepackage{soul} 

\usepackage{caption}
\usepackage{amsmath}

\DeclareMathOperator*{\argmin}{arg\,min}
\usepackage[ruled,vlined]{algorithm2e}
\usepackage{soul} 
\usepackage{array}
\usepackage{tabularx}

\usepackage{graphicx}
\usepackage[position=bottom]{subfig}

\acmConference[GECCO '21]{the Genetic and Evolutionary Computation Conference 2021}{July 10--14, 2021}{Lille, France}

\acmPrice{15.00}

\copyrightyear{2021}
\acmYear{2021}
\setcopyright{acmlicensed}\acmConference[GECCO '21]{2021 Genetic and Evolutionary Computation Conference}{July 10--14, 2021}{Lille, France}
\acmBooktitle{2021 Genetic and Evolutionary Computation Conference (GECCO '21), July 10--14, 2021, Lille, France}
\acmPrice{15.00}
\acmDOI{10.1145/3449639.3459303}
\acmISBN{978-1-4503-8350-9/21/07}

\begin{document}
\title{BR-NS: an Archive-less Approach to Novelty Search}

\author{Achkan Salehi}
\affiliation{%
  \institution{Sorbonne Université, CNRS, ISIR, F-75005}
  \city{Paris} 
  \state{France} 
}
\email{salehi@isir.upmc.fr}
\author{Alexandre Coninx}
\affiliation{%
  \institution{Sorbonne Université, CNRS, ISIR, F-75005}
  \city{Paris} 
  \state{France} 
}
\email{alexandre.coninx@sorbonne-universite.fr}
\author{Stephane Doncieux}
\affiliation{%
  \institution{Sorbonne Université, CNRS, ISIR, F-75005}
  \city{Paris} 
  \state{France} 
}
\email{stephane.doncieux@sorbonne-universite.fr}
%


\renewcommand{\shortauthors}{Salehi et al.}

\begin{abstract}
  As open-ended learning based on divergent search algorithms such as Novelty Search (NS) draws more and more attention from the research community, it is natural to expect that its application to increasingly complex real-world problems will require the exploration to operate in higher dimensional Behavior Spaces (BSs) which will not necessarily be Euclidean. Novelty Search traditionally relies on k-nearest neighbours search and an archive of previously visited behavior descriptors which are assumed to live in a Euclidean space. This is problematic because of a number of issues. On one hand, Euclidean distance and Nearest-neighbour search are known to behave differently and become less meaningful in high dimensional spaces. On the other hand, the archive has to be bounded since, memory considerations aside, the computational complexity of finding nearest neighbours in that archive grows linearithmically with its size. A sub-optimal bound can result in "cycling" in the behavior space, which inhibits the progress of the exploration. Furthermore, the performance of NS depends on a number of algorithmic choices and hyperparameters, such as the strategies to add or remove elements to the archive and the number of neighbours to use in k-nn search. In this paper, we discuss an alternative approach to novelty estimation, dubbed Behavior Recognition based Novelty Search (BR-NS), which does not require an archive, makes no assumption on the metrics that can be defined in the behavior space and does not rely on nearest neighbours search. We conduct experiments to gain insight into its feasibility and dynamics as well as potential advantages over archive-based NS in terms of time complexity.
\end{abstract}

%
%



\maketitle

\section{introduction}

Recent research seems to indicate that solutions to many fundamental and long-standing problems such as general artificial intelligence are likely to be reached through open-ended exploration of problems and solutions rather than manual engineering of different algorithmic components \cite{clune2019ai,stanley2019open, wang2019paired}. Such open-ended processes will require the divergent exploration of parameters in order to avoid deceptive minima in highly non-convex loss or fitness landscapes and encourage diversity in the set of possible solutions. Parameter spaces that need to be explored are more often than not high-dimensional, and many of their dimensions or subspaces can have little to no correlation with the tasks at hand. Therefore, it is desirable to limit the search to useful areas. Novelty Search (NS) \cite{lehman2011abandoning}, Surprise Search \cite{gravina2016surprise} and Curiosity Search \cite{cully2017quality} are among divergent methods that define a \textit{behavior space} as a proxy for conducting the search. Such a space can be either hand-engineered or learned \cite{cully2018hierarchical,cully2019autonomous,meyerson2016learning,paolo2020unsupervised}. In this paper, we focus on NS, but the proposed method can also be applied to other search approaches.

While most behavior spaces that are used for NS in the literature are very low-dimensional (usually $<8$), it is reasonable to expect that as NS is applied to increasingly complex domains, the need for higher dimensional behavior descriptors will arise. This is problematic since NS traditionally makes use of k-nearest neighbours search relative to an archive of previously visited individuals which are usually considered to lie in Euclidean space. First, it is well-known that nearest neighbour search in high-dimensional spaces is ill-defined as the ratio of the distances between the nearest and furthest neighbours to a query point approaches $1$ for a large number of point distributions \cite{aggarwal2001surprising, beyer1999nearest}. Second, the time complexity of nearest neighbours lookup in the archive is linearithmic in the archive's size (memory requirements also grow, but this is rarely an issue on modern hardware). As a result, the archive has to be bounded. A sub-optimal higher bound, as demonstrated in \cite{mouret2015illuminating}, will lead to "cycling" in behavior space, which prevents the exploration to reach novel areas of the behavior space. Additionally, while NS seems to be to some degree robust to the strategies chosen to add/remove elements to/from the archive as well as to the number of neighbours to use in the knn search \cite{gomes2015devising}, those hyperparameters still impact its performance and some attention to their tuning is necessary.

In this paper, we discuss an alternative approach to traditional NS, dubbed Behavior Recognition Novelty Search (BR-NS), which does not require an archive, makes no assumptions on the structure of the behavior space and does not rely on nearest neighbours search. As a result, it could be a more suitable alternative for operation in high-dimensional settings. Additionally, we propose formalisations and metrics to characterise and evaluate undesirable properties that can lead a novelty-based algorithm to cycle in behavior space.

Note that while a number of previous works have investigated archive-less approaches to NS (\textit{.e.g} \cite{gomes2015devising, urbano2014generalization, cully2017quality}), they still base novelty estimation on k-nearest neighbours search but with respect to the current population. Unsurprisingly, their results seem to indicate that archive-based novelty estimation is more reliable and encourages better exploration and more uniform coverage of the behavior space. In contrast to those works, we propose to replace knn-based estimation with a mechanism that is inspired by intrinsic motivation and curiosity \cite{aubret2019survey} in Reinforcement Learning, in particular as defined in \cite{burda2018exploration}. However, while those methods are concerned with exploration of novel observations/states, our aim is to search for novelty in the behavior space. 

For the reader's convenience, we briefly introduce Novelty Search in the next section (\S\ref{NS-sec}). We dedicate \S\ref{br-ns-sec} to the description of the proposed method and its properties. Formalisations and evaluation tools on the notion of cycling are presented in \S\ref{sec_cyc}. Finally, we demonstrate the feasibility of BR-NS and discuss its dymanics and advantages in \S\ref{sec_experim}. Closing discussions and remarks are the subject of section \S\ref{sec_concl}.

\section{Novelty Search}
\label{NS-sec}

We place ourselves in an evolutionary computation setting, where the optimisation is carried over $G$ generations. At each generation $g$, the current population $\mathcal{P}^g$ is mutated to produce a set of offsprings $\mathcal{O}^g$, and $\mu \triangleq |\mathcal{P}^g|$ elements are selected from $\mathcal{P}^g \cup \mathcal{O}^g$ according to a single or multiple objectives to form the following generation. To guide that selection, the NS algorithm defines a \textit{novelty objective} whose computation traditionally relies on an archive $\mathcal{A}^g$ consisting of a subset of previously visited individuals.

More precisely, the Novelty Search algorithm assumes the existence of a metric behavior space $\mathfrak{B}$. Given an individual $x$ at generation $g \in [0, G]$ in parameter (\textit{i.e.} genotype) space $\mathcal{X}$ and a reference set $\mathcal{R}^g \subset \mathcal{X}$, NS defines the \textit{novelty objective} at generation $g$ as

\begin{equation}
  \mathcal{N}^g(x)=\frac{1}{k}\sum_{i=0}^k d(\phi(x), \phi(u_i))
  \label{eq_ns}
\end{equation}

\noindent where $\phi: X \rightarrow \mathfrak{B}$ is a mapping from parameter space to behavior space, and where the $u_i \in \mathcal{R}^g$ are such that the $\phi(u_i)$ are the $k$ nearest neighbours of $\phi(x)$ in the set $\phi(\mathcal{R}^g)$. The notation $d(.)$ in equation \ref{eq_ns} denotes a metric function defined in $\mathfrak{B}$, which is usually assumed to be the $\ell^2$ Euclidean norm. The reference set $\mathcal{R}^g$ is most often defined as $\mathcal{R}^g=\mathcal{A}^g \cup \mathcal{P}^g$. 

This objective has been shown to be sufficent to encourage exploration and to outperform fitness based optimisation in problems with deceptive optima (\textit{e.g.} \cite{lehman2011abandoning}). Furthermore, recent works \cite{doncieux2019novelty} suggest that $\mathcal{A}$ asymptotically converges to a uniform sampling of $\mathfrak{B}$ when the latter is bounded. As in \cite{lehman2011evolving, cully2017quality}, it is possible to combine the novelty objective of equation \ref{eq_ns} with a fitness objective, $\textit{e.g.}$ using multi-objective algorithms such as NSGA-2 \cite{deb2002fast} in order to constrain the search to more useful individuals.

\begin{algorithm}
  \Large
  \KwIn{environement \textbf{Env}, random frozen encoder $\xi^*$, trainable encoder $\xi_1$, initial population \textbf{pop}, population size $\mu$, number of offsprings $\lambda$, task success criterion $T_s$}
  \KwOut{trained encoder $\xi_1$, set of solutions $\mathcal{S}$}
  $\mathcal{D}=\emptyset$ \#training dataset\\
  $\mathcal{S}=\emptyset$ \\
  \SetAlgoLined
  \SetKwProg{Fn}{Function}{}{end}
  \Fn{\textbf{step}(\textbf{$p$},\ \textbf{env},\ \textbf{dataset})}
  {
    \textbf{$p.fitness,\text{\ } p.behavior$}=\textbf{env.}\textit{evaluate}($p$)\\
        $\mathcal{E}_0=\xi^*(p.behavior)$\\
        $\mathcal{E}_1=\xi_1(p.behavior)$\\
        \textbf{$p.novelty$}=$dist(\mathcal{E}_0,\mathcal{E}_1)$\\
        \textbf{dataset} $\leftarrow$ \textbf{dataset} $\cup$ ($p.behavior$,$\mathcal{E}_0$)
  }
  \For{$p \in$ \textbf{pop}}
  {
    \textbf{step}(\textbf{$p$}, \textbf{Env}, $\mathcal{D}$)
  }
  \For{$g \in [0, G)$}
    {
      \textbf{offsprings} $\leftarrow$ \textit{generate\_new\_agents}(\textbf{pop, $\lambda$})\\
      \For {$p \in$ \textbf{pop} $\cup$ \textbf{offsprings}}
      {
        \If {$T_s(p.behavior)$}
        {
          $\mathcal{S} \leftarrow \mathcal{S} \cup p$
        }
      }
      \For{$p \in $ \textbf{offsprings}}
      {
        \textbf{step}(\textbf{$p$}, \textbf{Env}, $\mathcal{D}$)
      }

      \textbf{pop}=\textit{select\_most\_novel}(\textbf{offsprings} $\cup$ \textbf{pop})\\

      $\xi_1 \leftarrow$ \textit{train($\xi_1$,$\mathcal{D}$)}\\
      $\mathcal{D} \leftarrow \emptyset$ 

    }
  \caption{\Large{The BR-NS algorithm}}
  \label{algo_pseudo}
\end{algorithm}

\section{BR-NS}
\label{br-ns-sec}

NS is traditionally based on novelty estimates that are computed relative to a reference set $\mathcal{R}$ of individuals, using the expression given in equation \ref{eq_ns}. In contrast, the proposed approach does not rely on relative distances between individuals. Instead, it uses the heuristic presented in the following subsection to recognise behaviors that have been previously seen (hence the name Behavior Recognition based Novelty Search: BR-NS).

\subsection{Recognition of previously encountered behavior}

Our aim here is to define a heuristic to determine whether a behavior has been previously visited during exploration. To achieve this, we consider two encoders

\begin{equation}
  \begin{split}
    & \xi_0: \mathfrak{B} \times \mathbb{R}^{k_0} \rightarrow E  \\
    & (\boldsymbol{b}, \boldsymbol{w}) \mapsto \xi_0(\boldsymbol{b}; \boldsymbol{w})
  \end{split}
\end{equation}

\noindent and

\begin{equation}
  \begin{split}
    & \xi_1: \mathfrak{B} \times \mathbb{R}^{k_1} \rightarrow E  \\
    & (\boldsymbol{b}, \boldsymbol{w}) \mapsto \xi_1(\boldsymbol{b}; \boldsymbol{w})
  \end{split}
\end{equation}

\noindent which are respectively parametrised by weight vectors in $\mathbb{R}^{k_0}$ and $\mathbb{R}^{k_1}$. At initialisation, we sample a random vector $\boldsymbol{w_0} \in \mathbb{R}^{k_0}$ and define the random embedding

\begin{equation}
  \xi^* (\boldsymbol{b}) = \xi_0 (\boldsymbol{b}, \boldsymbol{w_0})
\end{equation}

\noindent which takes a behavior descriptor $b \in \mathfrak{B}$ to a random embedding space $E$. During execution, for each $\boldsymbol{b} \in \mathfrak{B}$, we minimise

\begin{equation}
  \mathcal{M}(\boldsymbol{b}, \boldsymbol{w})=||\xi^*(\boldsymbol{b}) - \xi_1(\boldsymbol{b}, \boldsymbol{w})||_2^2
\end{equation}

\noindent by optimising the parameters $\boldsymbol{w} \in \mathbb{R}^{k_1}$ of $\xi_1$, and we define the behavioral novelty of $\boldsymbol{b}$ as equal to $\mathcal{M}(\boldsymbol{b},\boldsymbol{w})$. The intuition is that if the behavior $\boldsymbol{b}$ has not been encountered yet, then the difference between the embeddings produced by the two encoders should be large. Otherwise, the network $\xi_1$ has already been trained to produce the target $\xi^*(\boldsymbol{b})$, and therefore the error $\mathcal{M}$ between the two mappings should be small. In practice, we consider an individual $\boldsymbol{x} \in X$ novel if $\mathcal{M}(\phi(\boldsymbol{x}), \boldsymbol{w})$ is superior to a threshold $t \in \mathbb{R}^+$.

As in the case of archive-based NS, the novelty presented here will depend on the generation $g$. The notation $\mathcal{M}^g$ will be used from this point forward in order to make that dependency explicit.

\subsection{Network architectures and initialisation}

A natural choice for both encoders is to encode them as vanilla feed-forward fully connected networks where the output of layer $l$ is given by

\begin{equation}
  \boldsymbol{y}^l = \sigma (\boldsymbol{V}^l \boldsymbol{y}^{l-1})
\end{equation}

\noindent where the matrix $V^l$ is the set of learnable weights of the $l-$th layer, and where $\sigma$ is a non-linear activation function. Note however that in order to avoid low $\mathcal{M}(\boldsymbol{b})$ values for behaviors that haven't been visited yet, it is desirable to ensure that 

\begin{equation}
  \mathbb{E}[||\xi^*(\boldsymbol{b}) - \xi_1(\boldsymbol{b}, \boldsymbol{w_{init}})||_2^2] > 0.
  \label{eq_expect}
\end{equation}

\noindent at initialisation. Assuming that the weights of both encoders are initialised with usual methods such as He initialisation \cite{he2015delving}, they will follow a Gaussian distribution $\mathcal{N}(0, s^2)$. In that case, using activation functions in the last layer can have undesirable side-effects\footnote{For example, if we use the $ReLU$ activation in the last layer, the ouputs will follow a rectified normal distribution and preserve their input's norm \cite{arpit2019benefits}. In that situation, it will become apparent from expanding the left-hand side of equation \ref{eq_expect} that while the expectation in that equation is indeed positive, it also depends on the norm of the input behavior vector $\boldsymbol{b}$, which is undesirable both for the definition of novelty and training stability.}. However, using only a dense layer without any activation as the last layer results in a Gaussian distribution of the outputs and a positive expectation in equation \ref{eq_expect}, which will be proportional to $s^2$.

In practice, we use shallow networks (typically $3$ to $5$ layers) and make $\xi_1$ slightly deeper than $\xi_0$ to ensure that it can easily overfit the latter in few iterations. As in \cite{burda2018exploration}, we have found that the leaky ReLU activation produces better results than other activations. To minimise the probability of mapping similar areas to the same embedding, we set $dim(E)= c \times dim(\mathfrak{B})$ where $c\geq 1$ is a constant factor. While at initialisation, the expectation of equation \ref{eq_expect} is indeed positive due to the previously discussed choices, we initially train $\xi_1$ for a few epochs (typically $\sim 15$) on batches that have been uniformly sampled from the behavior space in order to ensure that it diverges from $\xi_0$.

\begin{figure}[ht]
  \centering

  \subfloat[]{\label{figur:1}\includegraphics[width=35mm]{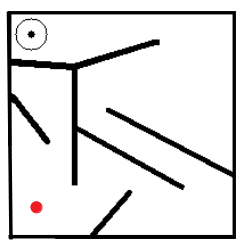}}
  \hspace*{0.2cm}
  \subfloat[]{\label{figur:2}\includegraphics[width=38mm]{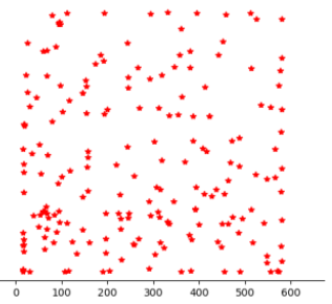}}
  \\
  \centering
  \subfloat[]{\label{figur:3}\includegraphics[width=77mm]{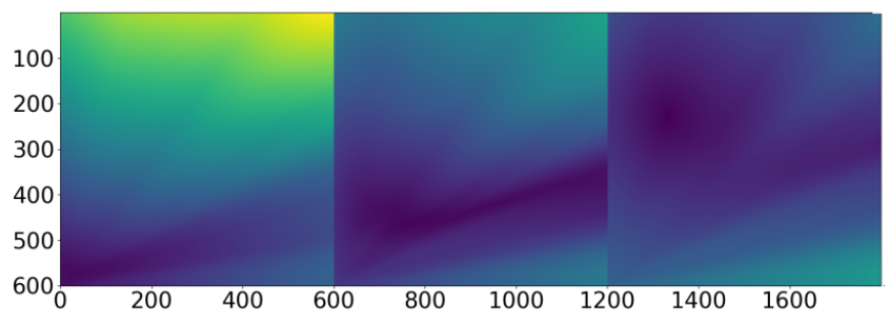}}
  \\
  \centering
  \subfloat[]{\label{figur:3}\includegraphics[width=72mm]{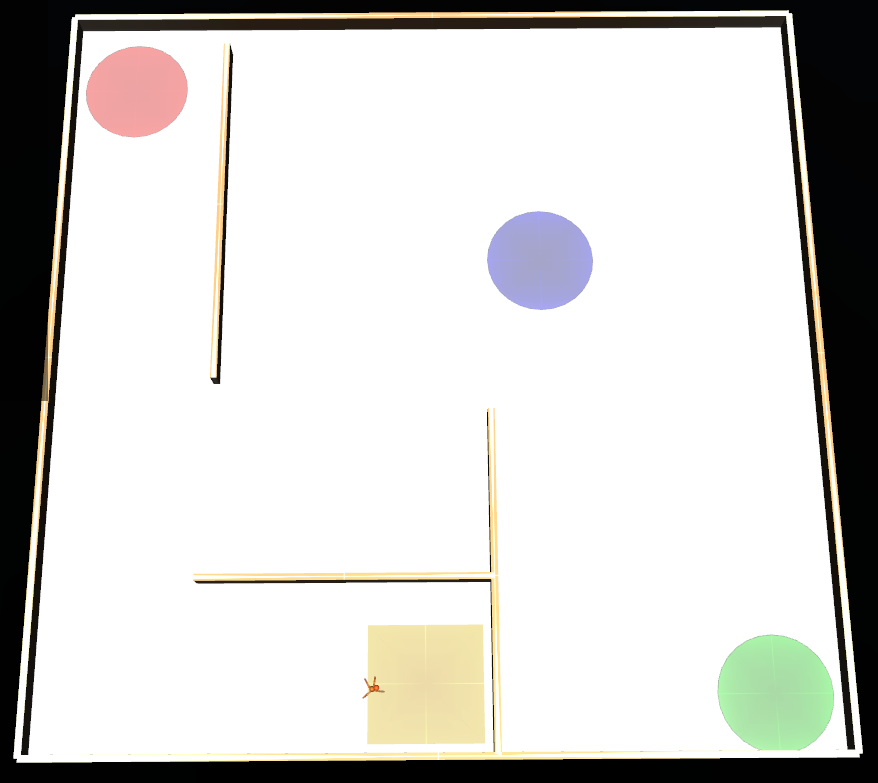}}
  \caption{\textbf{(a)} The deceptive maze environment, where the objective of the agent is to go from the red point to the goal area indicated by the small neighbourhood around the black point. \textbf{(b)} The archive in archive-based NS converges to a uniform sampling of the behavior space since it is bounded. \textbf{(c)} Example illustration of BR-NS' novelty at generations $5$, $100$, $150$ with a population size of $100$ (in the deceptive maze environment). Contrary to the archive-based method, here novelty itself (if expressed as a distribution) converges to uniformity. \textbf{(d)} The ant maze task, where the ant starts in the rectangular yellow area. The objective is to find paths such that the ant traverses all three red, green and blue regions.
  }
  \label{environment_figs}
\end{figure}

\begin{figure*}[h!t]

  \hspace*{-0.2cm}
  \subfloat[]{\label{figur:1}\includegraphics[width=67mm]{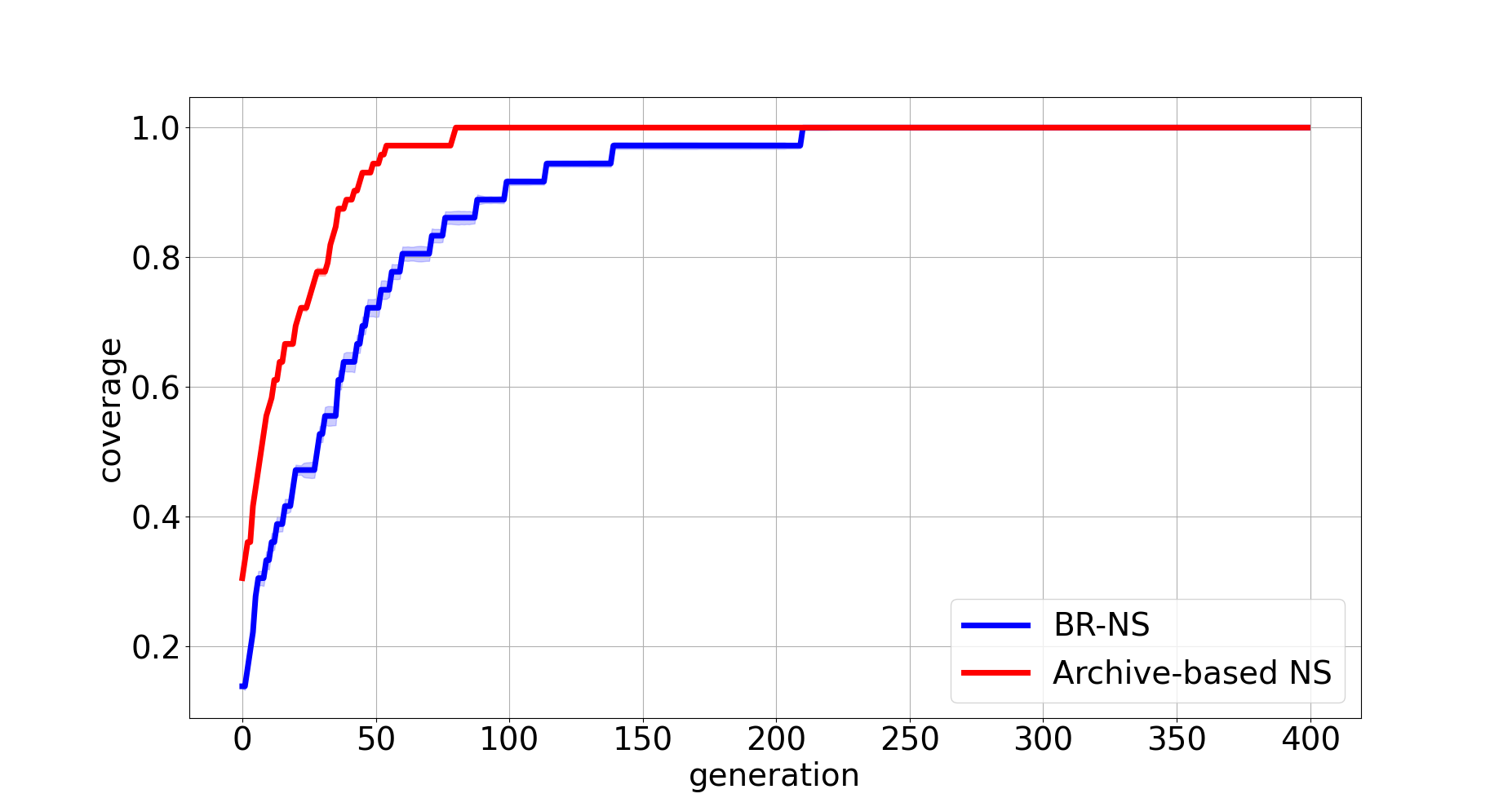}}
  \hspace*{-0.7cm}
  \subfloat[]{\label{figur:2}\includegraphics[width=67mm]{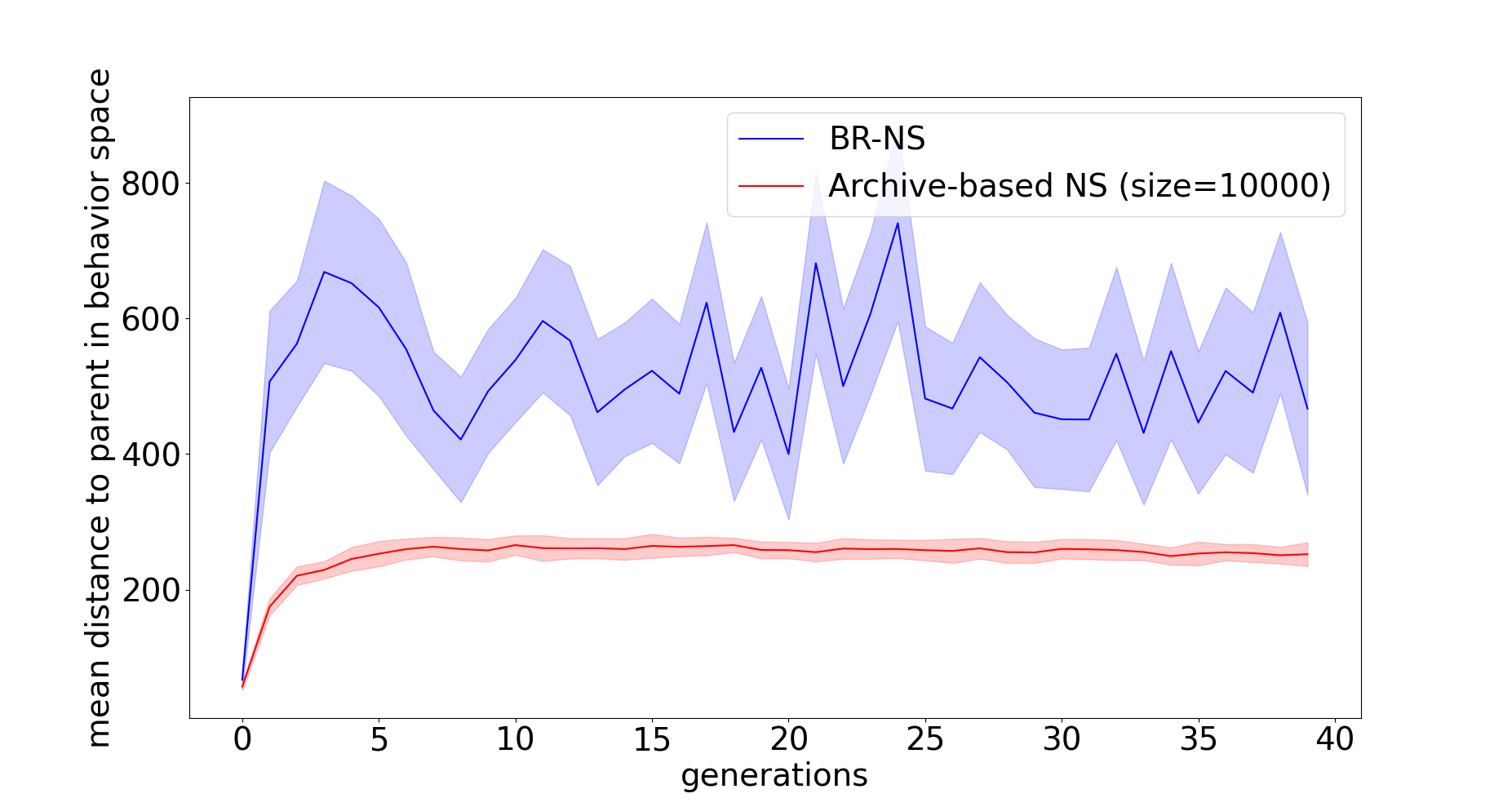}}
  \hspace*{-0.7cm}
  \subfloat[]{\label{figur:3}\includegraphics[width=67mm]{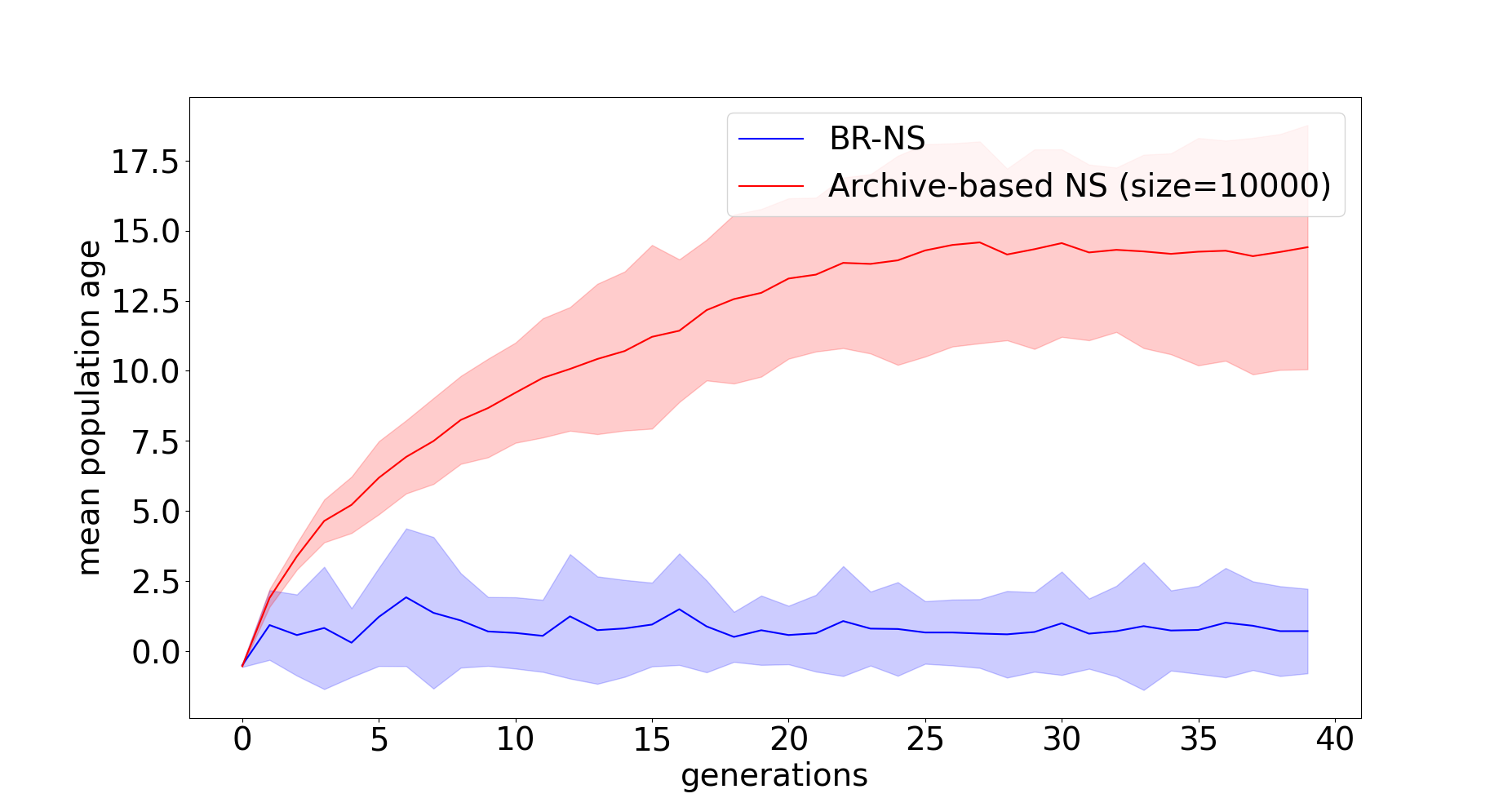}}
  \\
  \subfloat[]{\label{figur:1}\includegraphics[width=60mm]{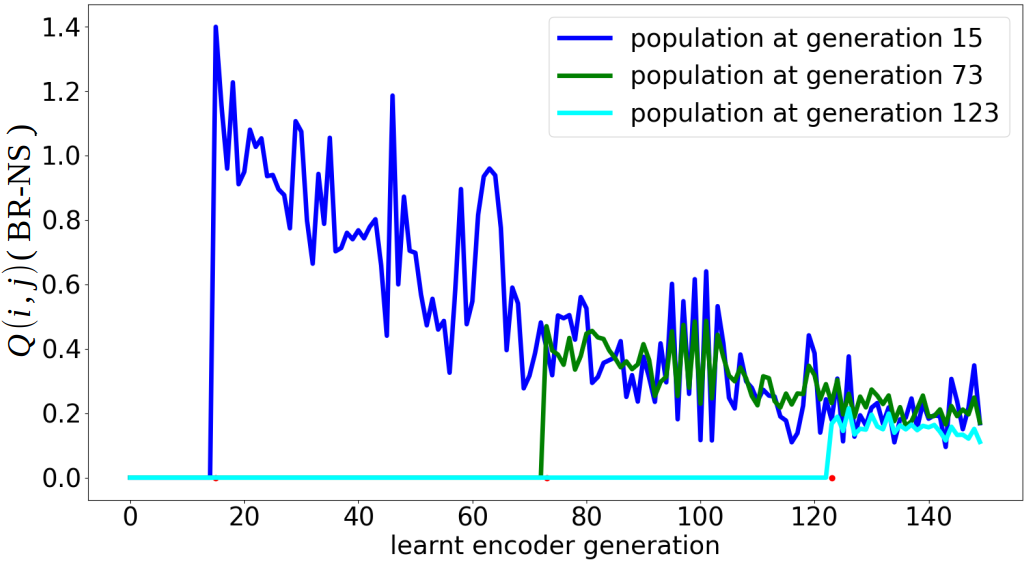}}
  \hspace*{+0.0cm}
  \subfloat[]{\label{figur:1}\includegraphics[width=60mm]{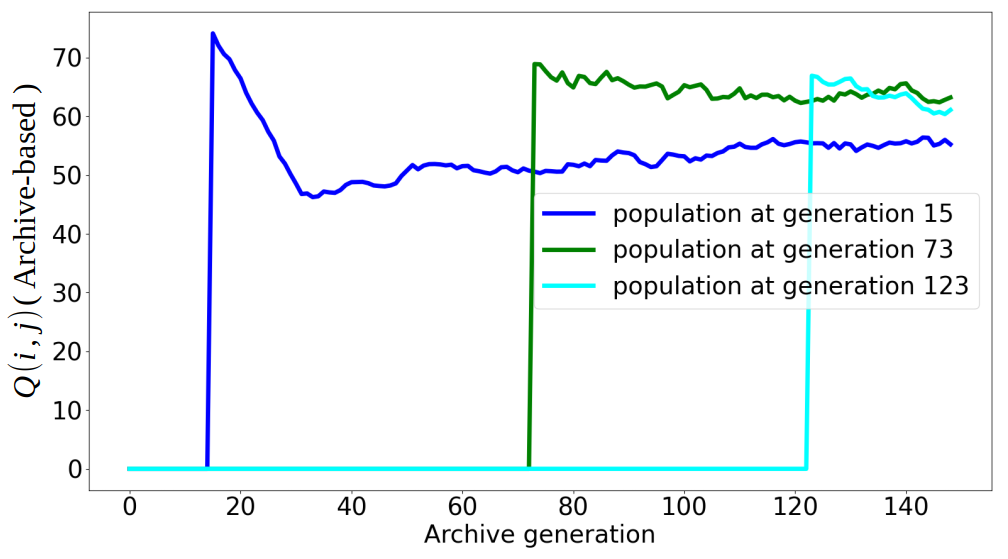}}
  \hspace*{-0.3cm}
  \subfloat[]{\label{figur:1}\includegraphics[width=69mm]{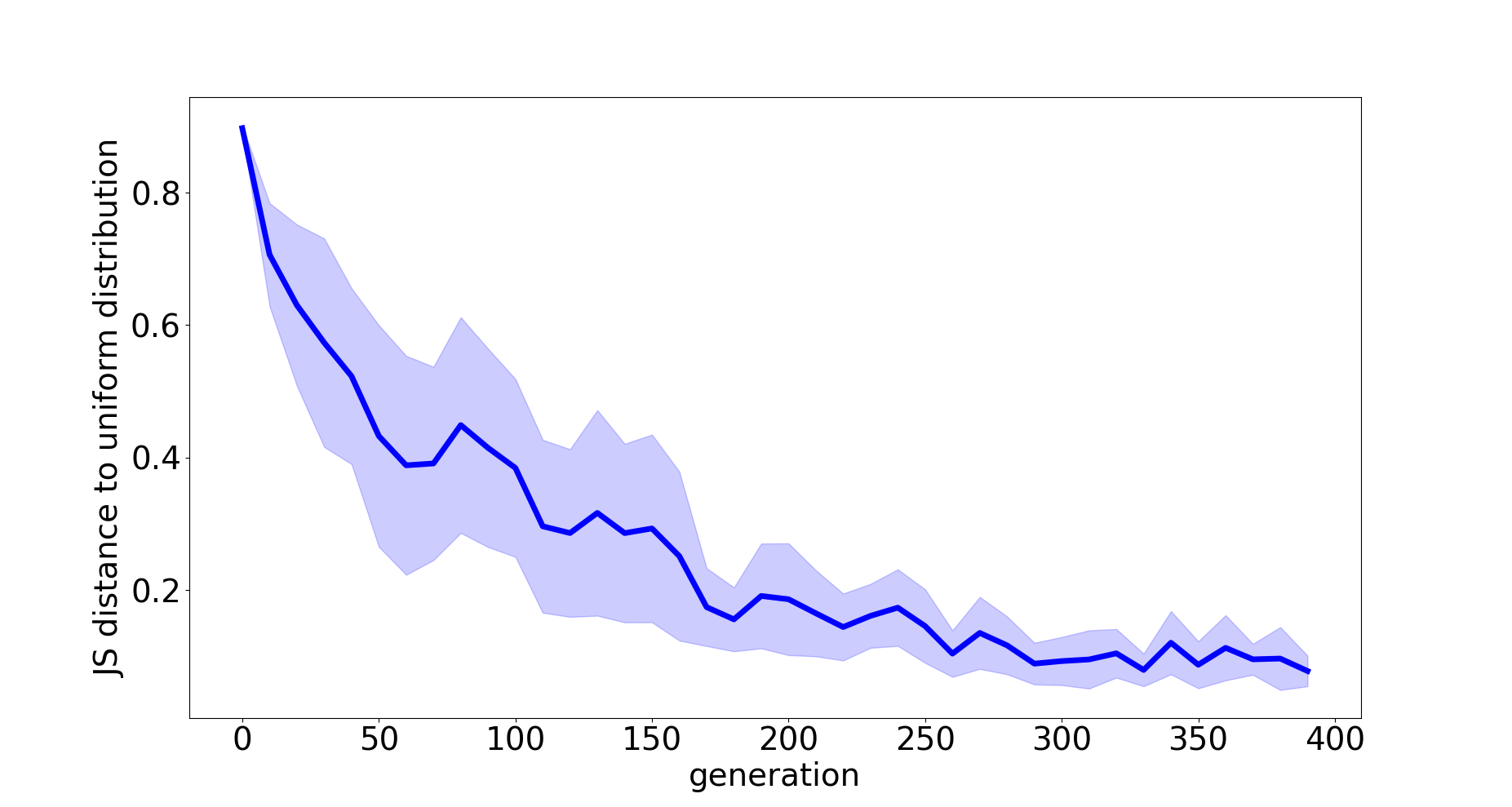}}
  \caption{Experimental results on the deceptive maze environment. Figure (d) illustrates $Q(i,j)$ values for three randomly selected generations $i$, for all $j>i$, based on the novelty provided by BR-NS. Figure (e) provides the same, except for archive-based NS.}
  \label{figure_deceptive_maze}
\end{figure*}

\subsection{Time complexity analysis}
\label{sec_time_comp}

In archive-based methods, the complexity of partitioning the behavior space and retrieving the $k$ nearset neighbours at a given generation is $O(Mlog(M))$ where $M$ is the cardinality of the reference set $\mathcal{R}$\cite{mouret2015illuminating, cormen2009introduction}. As a result, it will grow as execution progresses until the archive reaches its upper bound $M^{upper}$. In contrast, the complexity of the proposed method depends on population size and behavior space dimensionality, and therefore does not increase with time. As the computational costs of BR-NS are dominated by matrix multiplications in the encoders forward and backward passes, the method is of $O(Nd_m^2)$ complexity\footnote{Note that in general (denoting the number of epochs $n_{e}$ and network depth as $n_{d}$), that complexity is $O(n_{e}n_{d}Nd_m^2)$. However in our work, both $n_{e}$ and ${n_d}$ are negligible as they are both small constants ($\leq 5$) and therefore don't contribute to asympotic bounds.}  with $d_m=max(d_b,d_h, d_e)$ where $d_b, d_h, d_e$ respectively represent the dimensions of behavior descriptors, hidden layers and embeddings, and where $N$ denotes population size. Note that at each generation, $\xi_1$ is updated with few (typically $1$ to $5$) optimisation steps, and that the networks are both shallow. Also, in this analysis, we've assumed naive matrix multiplication, and multiplication algorithms that are implemented in modern libraries have lower computational requirements \cite{coppersmith1987matrix, alman2020refined}. In order to ensure sufficient coverage of a $d_m$ dimensional behavior space and avoid the cycling phenomon previously described, one has to choose $M^{upper}\gg d_m$, especially if the search space is unbounded. Note that in practice, it also holds that $N\ll M^{upper}$. This means that, as highlighted by our results in \S\ref{sec_experim}, it becomes preferable to use the proposed method in higher dimensional spaces once the archive size reaches a certain threshold.

\section{On cycling in behavior space}
\label{sec_cyc}

Given an ideal exploration algorithm, a key property that we expect it to hold would be its ability to keep track of the visited areas of behavior space in an optimal manner. An algorithm failing to do so is at risk of getting trapped in a cycle as it continually visits the same subspaces. This ability directly depends on the novelty estimator definition that is considered. Let us denote with $\Omega^g(\textbf{x})$ an arbitrary novelty function at generation $g$ (that is, $\Omega^g$ can be either $\mathcal{N}^g(\textbf(x))$, $\mathcal{M}^g(\textbf(x))$ or any other function used to drive exploration in NS). Furthermore, let us define

\begin{equation}
  Q(i,j)= \frac{1}{\mu}\sum_{x \in \mathcal{P}_i} \Omega^j(x) 
\end{equation}

\noindent as the mean novelty of the population at generation $i$ as estimated by the novelty function of generation $j$. Given generation $i$, We expect the following to hold for an ideal novelty function:

\begin{equation}
  \eta(i)\triangleq\mathop{\mathbb{E}}_{(j>i)}[\frac{Q(i,j)}{Q(i,i)}] < 1 
  \label{expectation_ratio}
\end{equation}

\noindent as we expect the novelty of previously visited areas to decrease with time. A novelty function for which the above does not hold can get trapped into cycling behaviors, similar to the case of archive-based novelty with a small archive \cite{mouret2015illuminating}. As the expectation of equation \ref{expectation_ratio} is sensitive to outliers and fluctuations as well as being insufficient for capturing longer-term tendencies in the evolution of novelty, we define a second, complementary measure as follows. Noting $i^*$ the generation at which the mean novelty of population $\mathcal{P}_i$ reaches its minimum, \textit{i.e.} $i^*=\argmin_j Q(i,j)$, we define

\begin{equation}
  \kappa(\mathcal{P}_i)= \sum_{(j>i^*)}\mathbb{I}(Q(i,i^*) + \epsilon < Q(i,j))
\end{equation}

\noindent where $\mathbb{I}$ is the predicate function and $\epsilon \in \mathbb{R}$ is a margin. Intuitively, $\kappa$ counts the number of significant deviations from the minimum novelty of $\mathcal{P}_i$ that happen after generation $i^*$. Low values for $\kappa$ will indicate that the algorithm correctly keeps track of previously visited areas, while large $\kappa$ values will indicate that previously visited areas and thus individuals are frequently becoming novel again. The margin $\epsilon$ is necessary to ignore fluctuations that are insignificant with respect to the size and dimensionality of the space based on which novelty is defined. In the case of archive-based NS, that space is the behavior space itself, while in the case of BR-NS, it is the embedding space defined by $\xi_0$. In this work, we have set $\epsilon=0.1median(\{d(\textbf{z}_i,\textbf{z}_j)\}_{(i,j)})$ where $\textbf{z}_i, \textbf{z}_j$ are vectors living in the appropriate space.

\section{Experiments}
\label{sec_experim}

The objective of the following experiments is to gain insight into those questions:

\begin{itemize}
  \item Is BR-NS feasible? If so, how does it compare in terms of dynamics with archive-based NS? 
  \item Are there situations in which BR-NS has a computational advantage compared to archive-based NS?
\end{itemize}

While we feel that the fact that BR-NS does not rely on $k$-nn search can have additional advantages beside computational ones because of the different behavior of $L_k$ norms in higher dimensions \cite{beyer1999nearest, aggarwal2001surprising}, we leave experimental investigations on that matter for futur works.

To answer the first question, we use the deceptive maze \cite{lehman2011abandoning} environment and present a study of the dynamics of NS in a spirit similar to that of \cite{doncieux2019novelty, doncieux2020novelty}. Furthermore, we show that both archive-based NS and BR-NS are well-behaved according to the criteria defined in the previous section. In the deceptive maze environment, illustrated in figure \ref{environment_figs} (a), a robot equipped with a few proximity sensors is positionned at the bottom left (red point in the figure) and its objective is to navigate the maze to reach the goal area on the top left (small neighbourhood around the black dot). As expected, distance based fitnesses result in agents that get trapped in the deceptive cul-de-sacs of the maze while NS effortlessy finds agents that reach the goal. As in \cite{lehman2011abandoning, doncieux2019novelty}, the behavior descriptor that we use for this task is the last $2d$ position of the agent. 

\begin{figure*}[h!t]
  \centering
  \subfloat[]{\label{figur:1}\includegraphics[width=64mm]{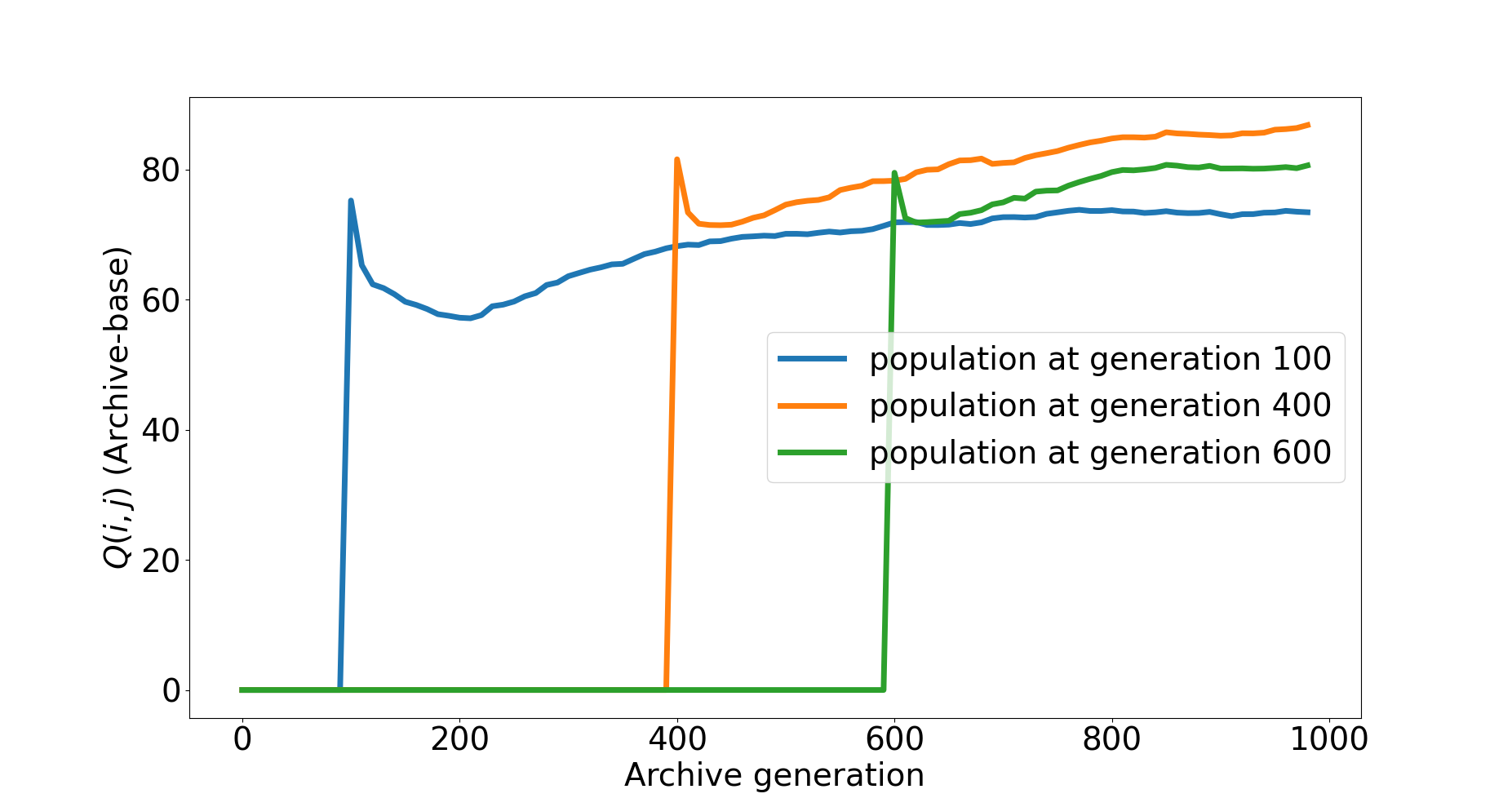}}
  \hspace*{-0.5cm}
  \subfloat[]{\label{figur:2}\includegraphics[width=64mm]{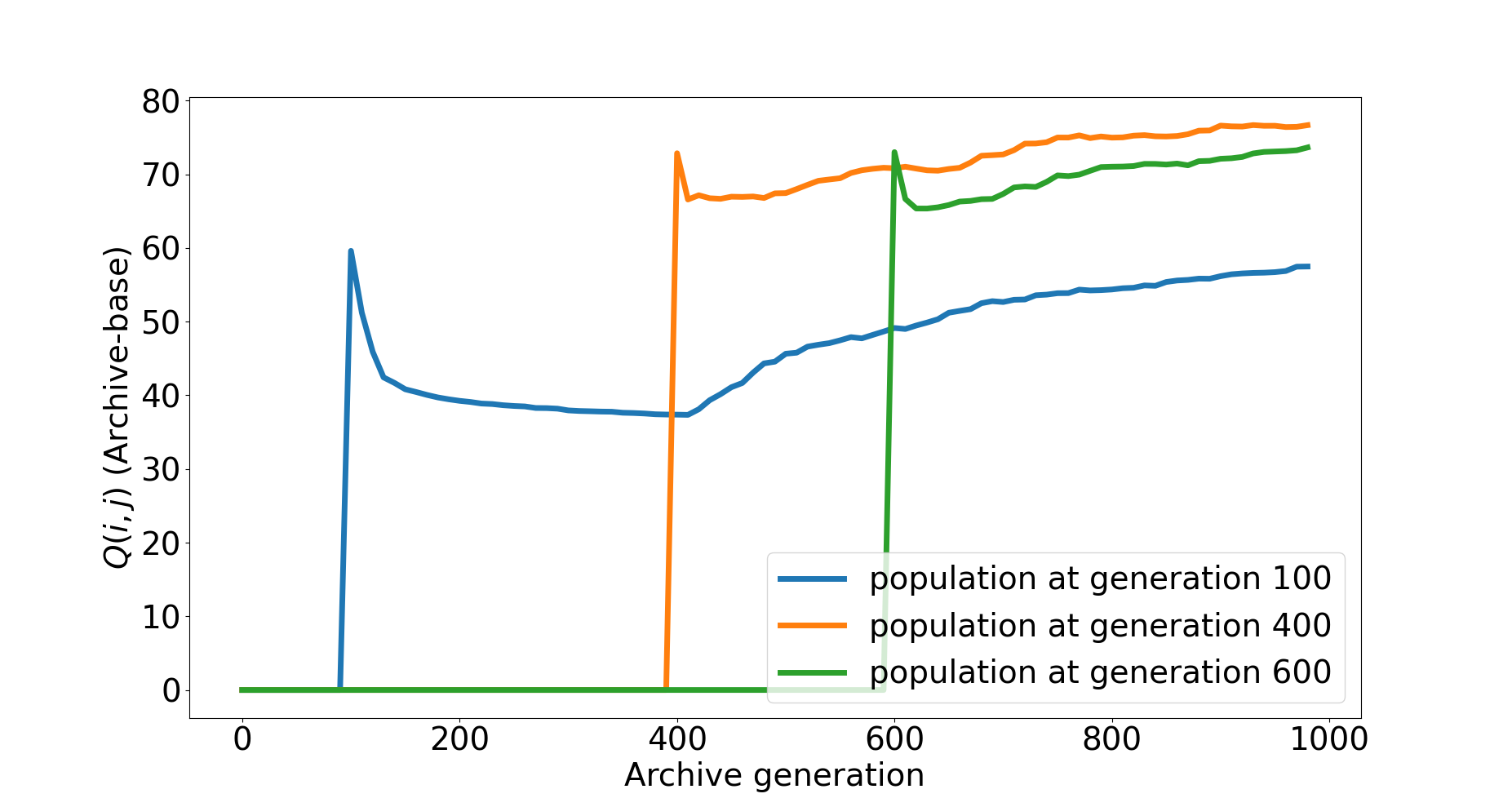}}
  \hspace*{-0.5cm}
  \subfloat[]{\label{figur:3}\includegraphics[width=64mm]{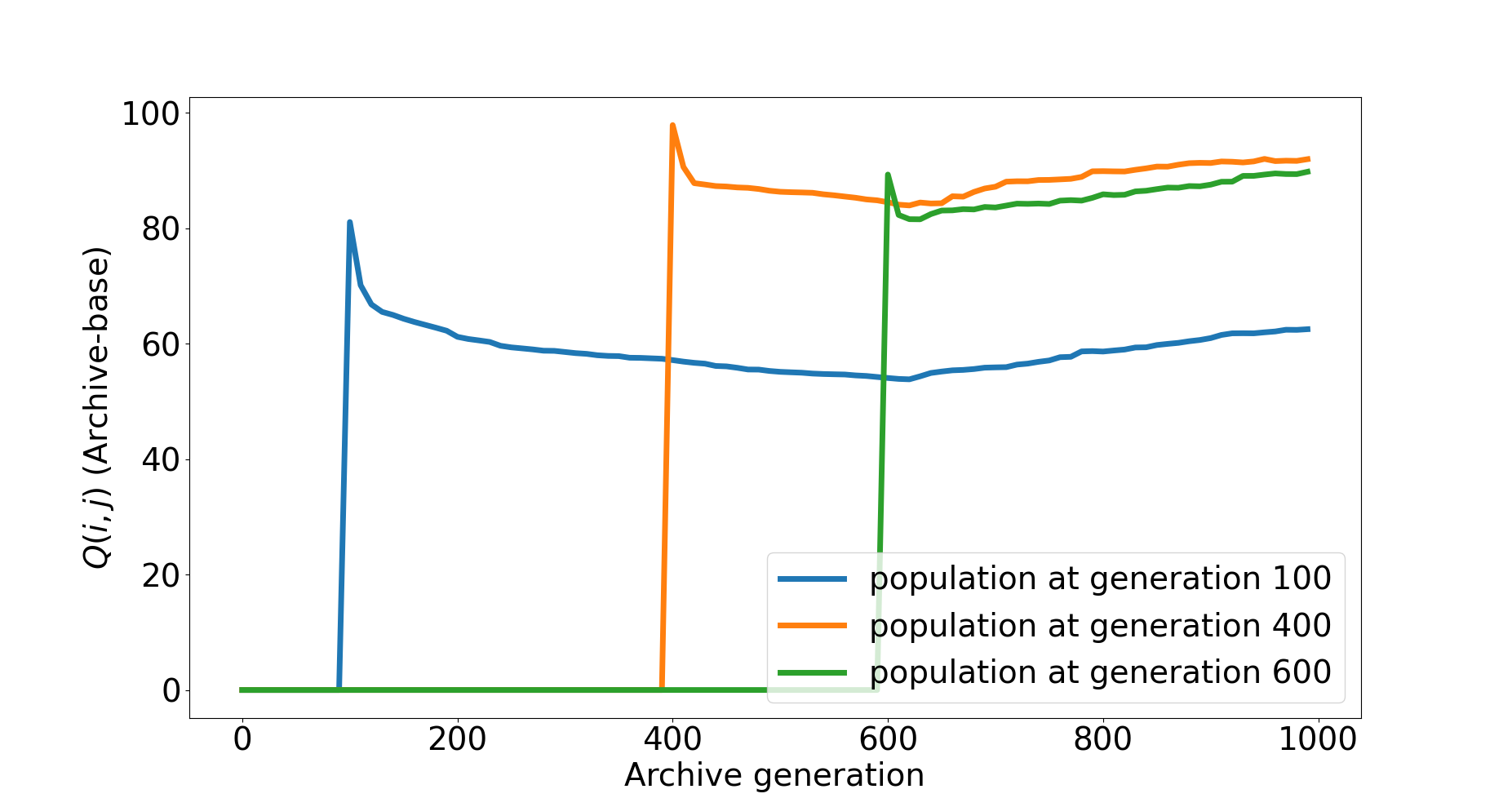}}
  \\
  \subfloat[]{\label{figur:4}\includegraphics[width=64mm]{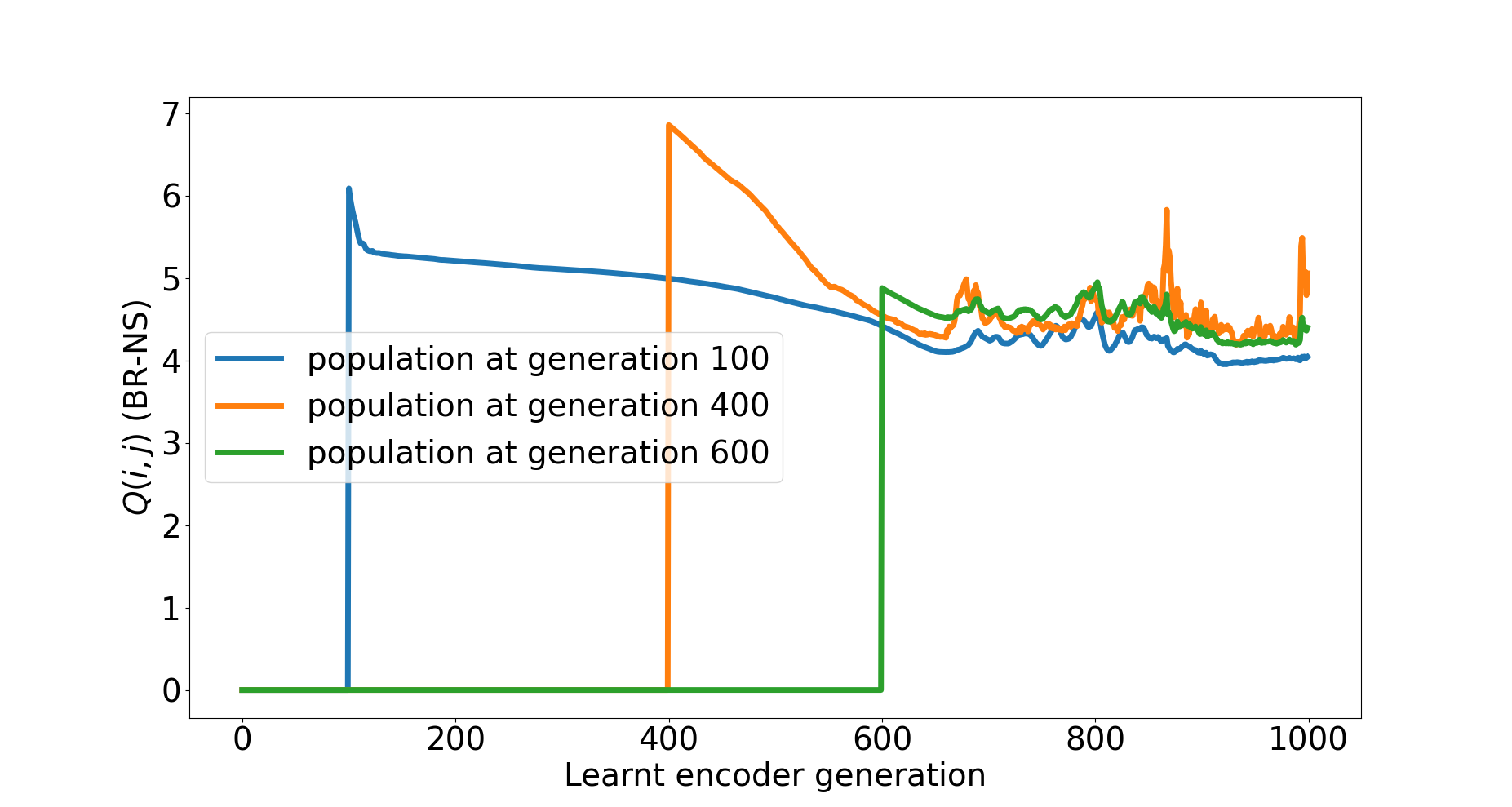}}
  \hspace*{-0.5cm}
  \subfloat[]{\label{figur:4}\includegraphics[width=64mm]{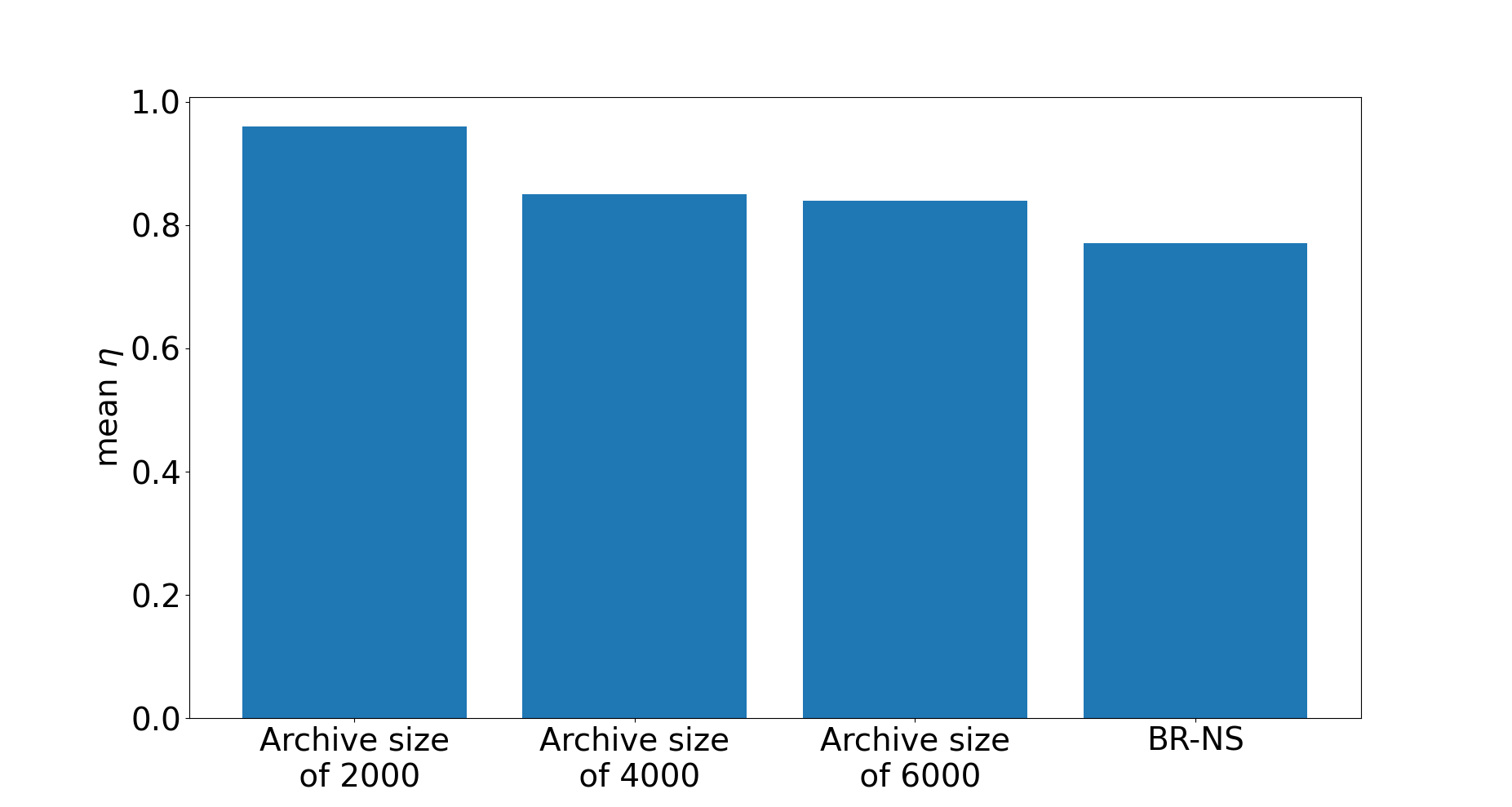}}
  \hspace*{-0.5cm}
  \subfloat[]{\label{figur:4}\includegraphics[width=64mm]{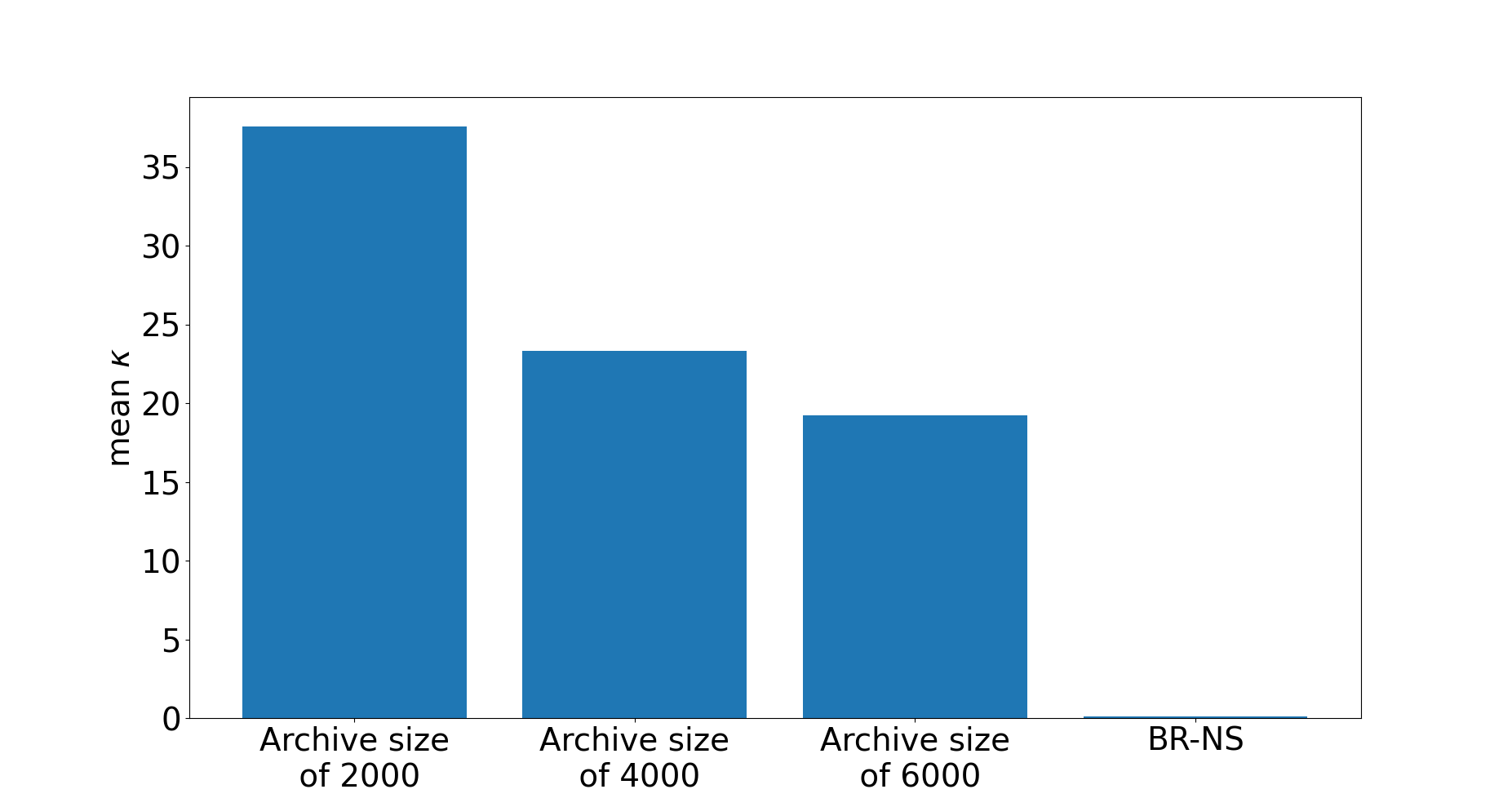}}
  \caption{Experimental results on the Ant environment. (a), (b), (c) and (d) all illustrates $Q(i,j)$ values for arbitrarily chosen values of $i$ with $j>i$. While (a), (b) and (c) correspond to archive-based NS with (respectively) archive sizes of $2000$, $4000$ and $6000$, (d) corresponds to BR-NS. (e), (f) provide the $\eta$ and $\kappa$ statistics.}
  \label{ant_results}
\end{figure*}

In order to answer the second question, we define a higher dimensional task (figure \ref{environment_figs} (d)) in an environment adapted from \cite{frans2017meta}. In that environment, the agent controls a four-legged "ant". The observation space is a subset of $\mathbb{R}^{29}$ (it is mainly composed of relative joint positions, angular velocities and accelerations), and the action space is $[0,1]^8$. We define the task as follows: starting in the yellow area of the large maze, the objective is to find paths such that the ant traverses all three of the blue, green and red areas shown on the map. We define the behavior space for this task as a $32$-dimensional space, where each point is the result of sampling the path traveled by the ant uniformly at $16$ locations. By defining this task, we firstly seek to highlight the cycling phenomenon that can happen in higher dimensional spaces if the archive used in archive-based NS is not large enough. Secondly, our aim is to show that in such cases, BR-NS can be a less compute-intensive alternative that retains similar exploration capacities to archive-based NS. 

In all experiments, we use simple fully connected feed-forward neural networks with $3$ layers, $10$ hidden units per layer and $tanh$ activations. Regarding the networks used in novelty computation for the presented method, both $\xi_0$ and $\xi_1$ map inputs of dimensionality $d_b$ to outputs of dimensionality $2*d_b$, have $3*d_b$ neurons per layer, and use leaky ReLU activations. While $\xi_0$ is $3$ layers deep, $\xi_1$ is composed of $5$ layers for all experiments. The latter is trained usign the ADAM optimiser. The learning rate is set to $1e-2$ in both environments. The mutation operator used in all experiments is the bounded polynomial operator \cite{deb2002fast}. 

An implementation of BR-NS is provided at \url{https://github.com/salehiac/BR-NS}.

\subsection{Deceptive maze results}

All results in this subsection are averaged over $20$ runs. In these experiments, $\lambda$ and $\mu$ are both set to $100$. The selection at each generation is elitist in terms of novelty. For Archive-based NS, the archive has a limit of $10000$ elements, with a growth rate of $6$ and a value of $k=15$ for $k$-nn search. As in \cite{doncieux2019novelty, doncieux2020novelty}, we evalute coverage and uniformity as well as two of the population properties that are associated with evolvability \cite{doncieux2020novelty}, namely the evolution of individual ages and the distances between offsprings and parents. The evolutions of coverages, which are evaluated based on a grid in similar fashion to \cite{doncieux2019novelty}, are illustrated in figure \ref{figure_deceptive_maze}(a) (The grid size is set to $6\times 6$). While both methods eventually reach full coverage of the behavior space, BR-NS is slightly slower to converge. This is however not surprising, for two reasons: first, the archive size and growth are rather large for such a low-dimensional and bounded behavior space. Second, the exploration of BR-NS is dictated by the embedding space, which does not not necessarily preserve the structure of the behavior space. Regarding uniformity, it is important to note that while in archive-based NS, the archive itself converges to a uniform sampling (as illustrated in figure \ref{environment_figs}(b)) \cite{doncieux2019novelty}, in the case of the proposed method, it is the novelty distribution itself that converges to a uniform density, as shown both qualitativally and quantitativally in (respectively) figures \ref{environment_figs}(c) and \ref{figure_deceptive_maze}(f). As for population dynamics, the evolution of parent to offspring dynamics and the evolution of mean individual ages (figure \ref{figure_deceptive_maze}(b) and (c)) indicate that the entire population is continuously being replaced with individuals that are far away in behavior space. Notice that the larger distance to parents in the case of BR-NS is due to the exploration being driven by novelty based on the embedding space defined by $\xi_0$, which as stated previously does not preserve the structure of the behavior space. While the coverage of archive-based NS progresses as an expansion in behavior space, the proposed method has a tendency to "jump" from place to place. 
We also verify if the compared approaches are well-behaved with respect to the statistics discussed in \S\ref{sec_cyc}. Figures \ref{figure_deceptive_maze}(d), (f) show $Q(i,j)$ values for three arbitrarily selected values of $i$ and all $j>i$ for those $i$. As it can be seen in these figures, the novelty of the population decreases with time. The inequality of equation \ref{expectation_ratio} is verified for all generations and $\kappa$ is close to zero for both methods, and no cycling occurs.

\subsection{Results on the Ant environment}

As evaluations on this task are costly, each experiment is performed only three times, and each simulation runs for $1500$ steps. To reduce the size of the search space, we associate a reward of $+1$ to each of the red, green and blue areas in the maze (figure \ref{environment_figs}(d)) that are given to the agent upon its first visit to one of those areas. We thus base the selection process on NSGA2 where we combine the novelty objective with the aforementioned fitness. To keep evaluation time reasonnable, population and offspring sizes are both set to $25$. For archive-based NS, we set the growth rate to $10$. The statistics $\eta$ and $\kappa$ are computed over the first $1000$ generations in order to focus on generation that might have been "forgotten" by the algorithm.

We first focus on archive-based NS with archive sizes of $2000$, $4000$ and $6000$ and on the computation of the statistics presented in \S\ref{sec_cyc}. For visual appreciation, example $Q(i,j)$ values are illustrated in figures \ref{ant_results} (a, b, c) (for three random $i$ and all $j>i$). It can be observed that in all situations, novelty seems to initially decrease before starting to increase again, even in some cases (the orange curve in figure \ref{ant_results} (b)) surpassing its initial values. Computation of the average $\eta$ and $\kappa$ statistics confirm these observations. As can be seen from figures \ref{ant_results} (e, f) all archive sizes verify the inequality of equation \ref{expectation_ratio}, but also produce high values for $\kappa$, indicating a clear trend in novelty increase as evolution progresses. All of the subfigures \ref{ant_results}(a, b, c, e, f) seem to support the hypothesis that a larger archive size results in better $\eta, \kappa$ statistics, and thus reduces the potential for cycling.

While as shown in figure \ref{ant_results} (d), the Novelty defined by BR-NS fluctuates to some degree (as expected due to the on-going training process to learn the outputs of $\xi^{*}$), both its $\kappa$ and $\eta$ values (figure \ref{ant_results}(e, f)) are lower than for archive-based NS and indicate that in general, previously visited behavior is not considered novel again.

\begin{figure}
\includegraphics[scale=0.17]{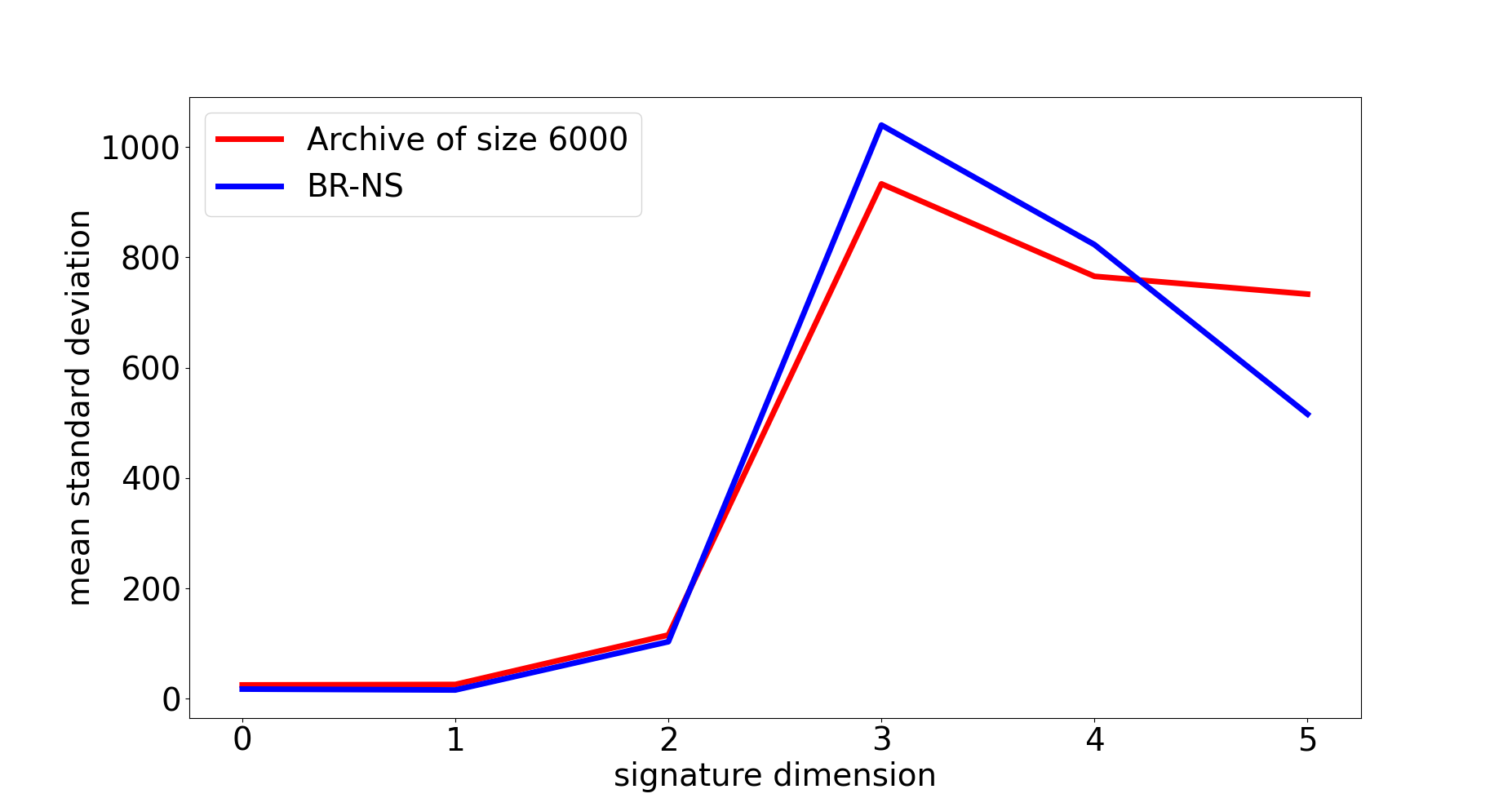}
  \caption{The standard deviation of found solutions along each dimension of their $6$-d signature.}
  \label{signatures_fig}
\end{figure}

We also compare the performance of the different methods in terms of exploration. To that end, we first note that all compared methods are able to quickly find at least one solution to the task within the first hundred generations. Second, we attempt to provide statistics on diversity, coverage and uniformity. However, their direct computation from the behavior space will not be meaningful (unless we run the experiments for an untractable number of generations) as the latter is a $32$-dimensional hypercube. We thus map each of the paths that the ant travels to a low dimensional feature space using the signature method  \cite{chevyrev2016primer, reizenstein2020algorithm}, truncated at six dimensions. The features produced via this method retain essential geometric properties from the path, that even if truncated at levels of four to six can provide sufficient information for character or drawing recognition \cite{fermanian2019embedding, yang2015character}. The plots in figure \ref{signatures_fig} show the standard deviation of the solutions generated with both BR-NS and archive-based NS (with an archive size of $6000$) along each of the dimensions of the truncated, $6$-d signature. Those results, which are averaged on all three experiments, indicate that the two methods seem to have similar reach in that feature space. In order to compute coverage and uniformity statistics, we define a grid of size $3\times 3 \times 5 \times 20 \times 15 \times 10$ (where the number of cells per dimensions is based on the magnitude of the standard deviation along that axis). Uniformity results, computed as the Jensen-shannon distance between each distribution, are reported in the second row of table I. In all cases, the Novelty distribution seems to converge to uniformity. While all methods have a coverage on the order of $1e-4$, it seems that the archive based method with an archive of size $6000$ has a slightly better coverage (table I, third row). This is coherent with our findings in the deceptive maze, where the convergence towards full coverage of the bounded space was slightly slower with BR-NS (figure \ref{figure_deceptive_maze} (a)). 

From a computational point of view, BR-NS, as expected from the theoretical discussion of section \S\ref{sec_time_comp}, has clear advantages over archive-based NS for all considered archive sizes (table I, first row). The timings reported in that table are the average time consumption of novelty computation for a single generation of size $25$.  

\vspace*{1.0cm}
\begin{center}
\begin{small}
  \captionof{table}{\textbf{}}
\begin{tabularx}{0.5\textwidth} {
  | >{\raggedright\arraybackslash}X
  | >{\raggedright\arraybackslash}X
  | >{\raggedright\arraybackslash}X
  | >{\raggedright\arraybackslash}X
  | >{\raggedright\arraybackslash}X | }
 \hline
  \ & Archive of size 2000 & Archive of size 4000 & Archive of size 6000 & BR-NS \\
 \hline
  mean execution time (ms)
  per novelty computation & 0.011  & 0.024  & 0.042 & \textbf{0.01}\\
\hline
  mean JS distance to uniform distribution &   0.16 & 0.12 & 0.18 & \textbf{0.05} \\
\hline
  mean coverage &   3e-4 & 4e-4 & \textbf{5e-4} & 3e-4 \\
\hline
\end{tabularx}
\end{small}
\end{center}
\vspace*{1.0cm}

\begin{figure}
  \center
  \includegraphics[scale=0.17]{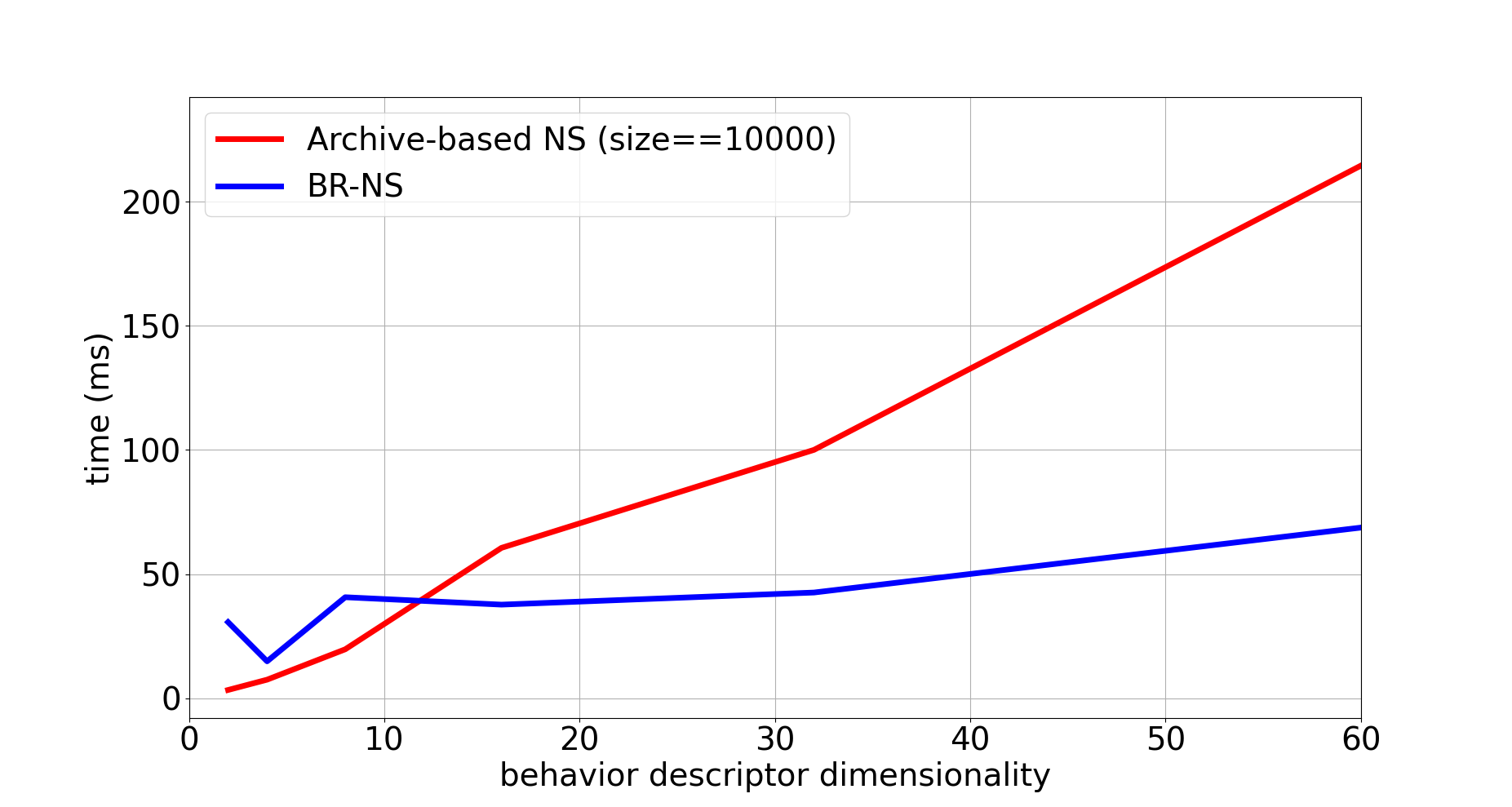}
  \caption{Evolution of time complexity of Novelty computation for a single generation as a function of behavior descriptor dimensionality, for BR-NS as well as archive-based NS using an archive size of $10000$. Note that this figure has been generated with $\lambda=\mu=100$.}
  \label{tc_evo}
\end{figure}


\subsection{Discussion} The experiments on the deceptive maze environment demonstrated the capacity of BR-NS to explore the behavior space, and the results on the Ant environment showed that this method can have a computational advantage when solving a given task once the archive size becomes too large. The task that we specified for the ant still was simple enough to be solved by archive based NS using relatively small archive sizes. As highlighted by figure \ref{tc_evo}, the computational gains brought by BR-NS will be more important in problems with larger and even higher-dimensional behavior spaces.

\section{Conclusion}
\label{sec_concl}

We presented an alternative to archive-based NS, which in contrast to the latter, does not rely on $k$-nn search, does not require an archive and does not make assumptions on the structure of the behavior space. Those properties make it an attractive candidate in large and higher dimensional spaces in which archive-based methods might present disadvantages related to computational complexity and $k$-nn search. The empirical investigations that were presented first demonstrated that the proposed approach exhibits exploration properties that is similar to archive-based NS on low dimensional and bounded spaces. Secondly, they highlighted its advantages in terms of computational complexity on a larger, high-dimensional task. 

\section{Acknowledgments}
This work has been supported by the FET project VeriDream\footnote{https://veridream.eu/}, that has received funding from the European Union's H2020-EU.1.2.2. research and innovation program under grant agreement No 951992.
\bibliographystyle{ACM-Reference-Format}
\bibliography{sample-bibliography} 

\end{document}